\documentclass{article}

\usepackage{microtype}
\usepackage{graphicx}
\usepackage{subcaption}
\usepackage{booktabs}
\usepackage{hyperref}
\usepackage{adjustbox}

\usepackage[preprint]{icml2026}

\usepackage{amsmath,amssymb,mathtools}
\usepackage{amsthm}

\usepackage[capitalize,noabbrev]{cleveref}

\begin{document}

\twocolumn[
    \icmltitle{Lngram: N-gram Conditional Memory in Latent Space}

    \begin{icmlauthorlist}
      \icmlauthor{Yunao Zheng}{bupt,liauto}
      \icmlauthor{Guoyang Xia}{bupt,liauto}
      \icmlauthor{Xiaojie Wang}{bupt}
      \icmlauthor{Lei Ren\textsuperscript{*}}{liauto}
    \end{icmlauthorlist}
    
    \icmlaffiliation{bupt}{Beijing University of Posts and Telecommunications (BUPT), Beijing, China}
    \icmlaffiliation{liauto}{Li Auto Inc., Beijing, China}
    
    \icmlcorrespondingauthor{Xiaojie Wang}{xjwang@bupt.edu.cn}

    \icmlkeywords{Lngram, N-gram, Latent Space}
    
    \vskip 0.3in
]

\begingroup
\renewcommand{\thefootnote}{}
\footnotetext{\hspace{-1.8em}\textsuperscript{*} Project Leader: Lei Ren.}
\endgroup

\printAffiliationsAndNotice{}

\begin{abstract}

Sequence modeling requires both compositional reasoning and local static knowledge retrieval, yet standard Transformers handle both through dense computation. Engram partially decouples retrieval from the backbone, but its token-based keys remain tied to text tokenization and hash compression. We propose Lngram, a latent-space conditional memory module that learns discrete symbols directly from hidden states and performs N-gram lookup over these symbols. This design removes the dependence on tokenizer IDs and naturally extends to non-text modalities. In our evaluated settings, Lngram outperforms Transformer and Engram baselines, consistently reduces perplexity in long-context language modeling, and effectively injects domain knowledge when added post hoc to pretrained models. Joint training with the backbone further surpasses full fine-tuning, while experiments on vision-language and vision-language-action tasks show overall gains. Analyses with LogitLens and CKA suggest that Lngram enables prediction-relevant information to emerge earlier, increasing effective depth with limited inference and memory overhead. Code is available at https://github.com/zyaaa-ux/Lngram.

\end{abstract}

\section{Introduction}

Transformer architectures have driven the development of multimodal models and have become the backbone of today’s large-scale models\citep{vaswani2017attention,devlin2019bert,brown2020language,dosovitskiy2021image,radford2021learning,alayrac2022flamingo}.A key reason for their success is that the unified attention--feedforward stack can handle both local patterns and global dependencies within a single framework\citep{vaswani2017attention}.From the perspective of computational function, however, standard Transformers still represent operations of fundamentally different natures using the same dense neural computation.Specifically, sequence modeling typically involves two kinds of subtasks: one is compositional reasoning, which requires deeper dynamic computation that changes with context; the other is knowledge retrieval, which depends primarily on matching local static patterns and is better implemented through low-cost lookup operations.Because standard Transformers lack a native lookup primitive, such retrieval can only be approximated through multiple layers of attention and feedforward networks.For example, to recognize a common multi-token entity, the model often has to gradually aggregate local context and complete the match in the early layers.Functionally, this process is closer to table lookup than to deep reasoning, and therefore consumes effective depth that could otherwise be used for subsequent compositional computation, ultimately affecting the model’s reasoning performance.

To separate this type of static retrieval from backbone computation, \citet{cheng2026engram} explored an alternative called Engram.Engram performs local suffix N-gram retrieval at designated layers: it first compresses tokenizer outputs into normalized identifiers, then maps them through deterministic multi-head hashing to several embedding tables, retrieves static vectors, and fuses them with the current hidden states through context-dependent gating and lightweight convolutions.However, this separation is still incomplete in Engram, as its retrieval keys are constructed entirely from tokenizer IDs and thus remain tied to the tokenizer’s segmentation scheme.Meanwhile, because the combinatorial space of N-grams grows rapidly with vocabulary size and order, practical implementations can only rely on hash-based compression to map a large number of patterns to a limited number of table entries, making collisions unavoidable.Moreover, the hash functions themselves are not learnable, making it difficult to adaptively correct matching errors according to the data and task.More importantly, models in non-text modalities often do not have a stable text tokenizer at all; vision and multimodal models typically rely on image patches, visual encoder features, or cross-modal connector modules rather than fixed text subword IDs.These limitations make such approaches better suited to experimental exploration than to direct use in real systems.

Therefore, if conditional memory is to be extended to more general representation spaces, the key is to learn discrete keys from hidden states rather than directly reusing tokenizer IDs.A GitHub demo of ROSA inspired us: by mapping hidden states to binary or low-bit routing codes through learnable projections and using them to construct keys for local conditional matching, it is possible to preserve a large portion of the information in the original hidden states\citep{blinkdl2026rwkvv8rosa}.Motivated by this idea, we replaced the original vectors in the key-value (KV) cache used for attention computation with binarized discrete representations in Qwen3 \citep{yang2025qwen3}.Experimental results show that, under windowed attention alone, the model’s performance on simple long-context tasks already approaches that of a global-attention baseline.Building on this observation, we adapt the discrete-hidden-state approach to the classical n-gram matching framework\citep{brants2007large}, and combine it with modern refinements to propose Lngram (Latent n-gram), a latent-space conditional matching mechanism.

Specifically, Lngram first discretizes the input hidden states into routing codes and constructs n-gram indices along the temporal dimension; it then performs exact table lookup based on these indices to retrieve the corresponding memory representations.The retrieved results are modulated by a contextual gating mechanism and added back to the original hidden states as a residual, after which they are fed into the original attention module.To address the non-differentiability introduced by hard discrete routing, we further design optional approximate and exact gradient backpropagation methods to ensure stable training.

We first evaluate Lngram on natural language processing (NLP) tasks.Under the same parameter count and training conditions, models equipped with Lngram outperform the baseline on all evaluation items.Subsequent experiments on vision-language models (VLMs) and vision-language-action models (VLAs) show that this gain is not limited to the text modality.LogitLens and CKA analyses show that Lngram reduces the need for the backbone to reconstruct static knowledge in the early layers, thereby increasing the effective depth available for compositional computation;at the same time, some local dependencies are handled by lookup operations, allowing the attention module to devote more capacity to global context modeling, which correspondingly improves performance in long-context scenarios.We also integrate Lngram as an additional component into pretrained models and train only the Lngram component on domain-specific data.The results show that this approach can effectively inject domain-specific knowledge, with performance close to that of full fine-tuning; when Lngram is jointly trained with the base model parameters, its performance is substantially better than that of full fine-tuning alone.

Finally, we evaluate the runtime and memory overhead of Lngram during the prefilling stage (prefill) and the autoregressive decoding stage (decode).Lngram’s online computation involves only a small number of linear layers and the table entries that are actually hit, and the table parameters can also be deployed separately from GPU memory in host memory.Therefore, even when the tables are large, the additional impact on throughput and memory usage remains small.Overall, Lngram shows that rewriting a class of local static matching operations originally performed by dense computation as conditional lookup can improve a model’s effective depth at low system cost, thereby enhancing reasoning performance.At the same time, this idea also provides a scalable retrieval primitive for multimodal models.

In summary, this paper makes the following contributions:
\begin{enumerate}

    \item We propose Lngram, a latent-space conditional matching mechanism that learns discrete routing keys from hidden states, constructs local n-gram indices in representation space, and retrieves memory vectors through exact table lookup before attention computation.

    \item We develop practical training mechanisms for hard discrete routing, including optional approximate and exact gradient propagation strategies, enabling Lngram to be trained stably together with standard Transformer backbones.

    \item We empirically validate Lngram across NLP, VLM, and VLA settings, showing consistent average improvements, improved use of effective depth, stronger long-context behavior, and efficient domain-specific knowledge injection with low runtime and memory overhead.
\end{enumerate}

\section{Architecture}

\subsection{Overview}

\begin{figure*}[t]
\centering
\includegraphics[width=0.95\linewidth]{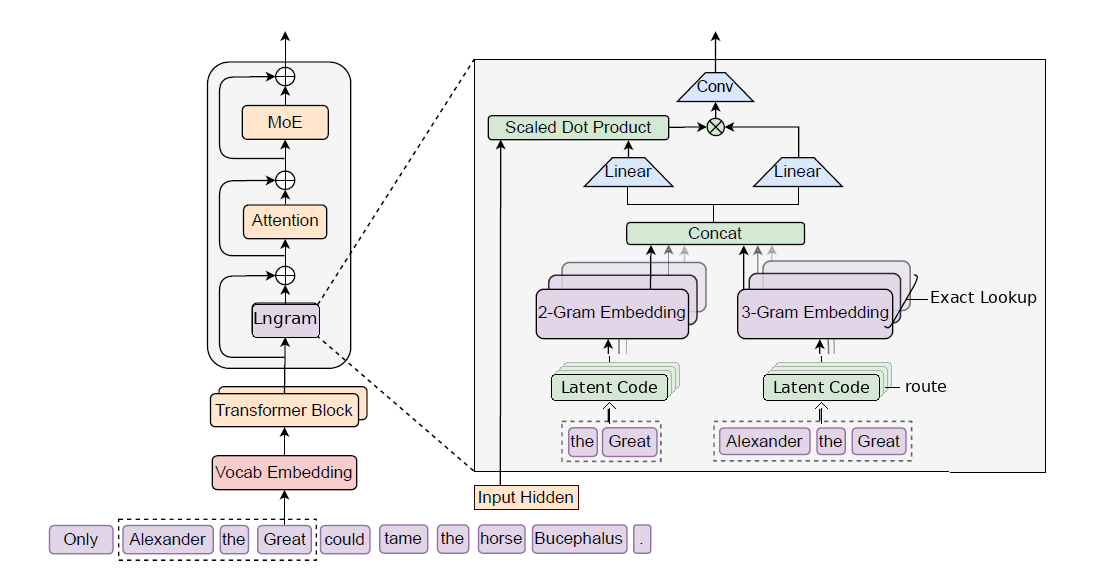}
\caption{The main architecture of Lngram.}
\label{fig:lngram}
\end{figure*}

As shown in Figure~\ref{fig:lngram}, Lngram is a conditional memory branch inserted into Transformer decoder layers, with the goal of separating the matching and storage of local static patterns from dense computation in the backbone network. For ease of exposition, we omit the batch dimension in what follows and denote the input to layer $\ell$ as
\[
H^{(\ell)} = [h_1,\ldots,h_T]^\top \in \mathbb{R}^{T\times d}.
\]

For each position $t$, Lngram proceeds through three stages: discretization, retrieval, and readout. First, the module maps the hidden state at the current position into multiple discrete symbols. It then uses these symbols to construct an exact $n$-gram key ending at position $t$ and retrieves the corresponding vectors from one or more memory tables. Finally, the current hidden state performs context-aware filtering and fusion over the retrieved results, which are then further refined by a short-range causal convolution. The resulting output is added back to the layer input as a residual branch, after which it is passed to the base layer's original self-attention and feedforward network. Because the addressing path contains hard discrete operations, during training we use counterfactual surrogate gradients to provide gradients for the discretization projection, while the table parameters and readout parameters are still updated in the standard way.

\subsection{Binary Discretization and Multi-Route Symbol Streams}

Because Lngram performs addressing in a discrete symbol space, it must first map continuous hidden states into discrete symbols. To avoid interfering with the base attention parameters, we introduce an independent discretization projection for Lngram:
\[
U = \mathrm{RMSNorm}(H), \qquad Z = U W_q,
\]
where $W_q \in \mathbb{R}^{d\times d}$ is a trainable parameter, and $Z=[z_{t,c}] \in \mathbb{R}^{T\times d}$ denotes the discretization logits.

We then apply hard-threshold binarization to each dimension:
\[
b_{t,c} = \mathbb{I}[z_{t,c} > 0].
\]

We partition the $d$ channels into routes of $M$ dimensions each, yielding
\[
R = \frac{d}{M}, \qquad c \leftrightarrow (r,j),
\]
where $r\in\{0,\ldots,R-1\}$ denotes the route index and $j\in\{0,\ldots,M-1\}$ denotes the bit index within that route. The symbol space size of each route is
\[
K = 2^M.
\]

For route $r$, we pack its $M$ bits into an integer symbol:
\[
a_{t,r} = \sum_{j=0}^{M-1} b_{t,(r,j)} 2^j,
\qquad
a_{t,r}\in\{0,\ldots,K-1\}.
\]

As a result, a hidden-state sequence of length $T$ is represented as $R$ parallel discrete symbol streams:
\[
\{a_{1,r}, a_{2,r}, \ldots, a_{T,r}\}_{r=0}^{R-1}.
\]

This construction does not rely on a tokenizer and confines subsequent matching to a controllable discrete space.

\subsection{Exact n-gram Retrieval}

Let $\mathcal{N}$ denote the set of $n$-gram orders used. For each $n\in\mathcal{N}$, we define an exact route-partitioned $n$-gram memory table
\[
E^{(n)} \in \mathbb{R}^{R K^n \times d_m},
\]
where $d_m$ is the continuous vector dimension of each table entry.

For position $t$ and route $r$, when $t\ge n$, the table address corresponding to the $n$-gram ending at position $t$ is
\[
g_{t,r}^{(n)} =
rK^n + \sum_{i=0}^{n-1} a_{t-n+1+i,r} K^i.
\]

The first term $rK^n$ distinguishes different routes, allowing all routes of the same order to share a single physical table without address collisions. The retrieved memory vector on that route is
\[
m_{t,r}^{(n)} = E^{(n)}[g_{t,r}^{(n)}] \in \mathbb{R}^{d_m}.
\]

Concatenating the results from all routes gives the retrieval result of that order at position $t$:
\[
e_t^{(n)} =
\mathrm{Concat}_{r=0}^{R-1} m_{t,r}^{(n)}
\in \mathbb{R}^{R d_m}.
\]

When $t<n$, the current prefix is insufficient to form a complete $n$-gram, and we set $e_t^{(n)}=0$.

In the single-table version, each order has only one table, and the number of distinguishable keys is fixed at $R K^n$. Under fixed $M$ and $n$, increasing $d_m$ can only increase the representation dimension of each table entry and cannot increase the number of distinguishable keys. Therefore, if one wishes to further increase memory capacity without changing the code space, a multi-table version of Lngram can be introduced.

Let the number of subtables be $S$. For each subtable $s\in\{1,\ldots,S\}$, we introduce an independent addressing projection $W_q^{(s)}$ and an independent table group $\{E_s^{(n)}\}_{n\in\mathcal{N}}$:
\[
Z^{(s)} = U W_q^{(s)},
\qquad
a_{t,r}^{(s)} =
\sum_{j=0}^{M-1} \mathbb{I}[z_{t,(r,j)}^{(s)} > 0] 2^j.
\]

The corresponding address and retrieval result are
\[
g_{t,r,s}^{(n)}
=
rK^n + \sum_{i=0}^{n-1} a_{t-n+1+i,r}^{(s)} K^i,
\]
\[
e_{t,s}^{(n)}
=
\mathrm{Concat}_{r=0}^{R-1} E_s^{(n)}[g_{t,r,s}^{(n)}].
\]

In the multi-table version, different subtables have different addressing projections and different table parameters, so the same hidden state can generate multiple parallel retrieval branches. In practice, the multi-table version can perform retrieval and normalization in subtable-wise blocks, thereby avoiding the need to materialize all branch outputs simultaneously.

\subsection{Context-Aware Readout}

The retrieved $e_t^{(n)}$ or $e_{t,s}^{(n)}$ is static memory independent of the current context. To mitigate the impact of mismatched retrievals on the backbone network, Lngram follows Engram in introducing a context-aware filtering mechanism at the readout stage.

We first consider the single-table version. For each order $n$, the retrieval result is projected into the Key/Value space:
\[
k_t^{(n)} = W_K e_t^{(n)} + b_K,
\qquad
v_t^{(n)} = W_V e_t^{(n)} + b_V.
\]

Here, $W_K$ and $W_V$ are shared across different orders. The hidden state $h_t$ at the current position already contains contextual information aggregated by the preceding layers, and we therefore use it as a dynamic query. After RMSNorm, we define the gate value of the $n$-th branch as
\[
\alpha_t^{(n)} =
\sigma\!\left(
\frac{
\mathrm{RMSNorm}(h_t)^\top \mathrm{RMSNorm}(k_t^{(n)})
}{
\sqrt{d}
}
\right).
\]

The final readout is
\[
v_t = \sum_{n\in\mathcal{N}} \alpha_t^{(n)} v_t^{(n)}.
\]

If a retrieval branch is inconsistent with the current context at position $t$, its similarity decreases and its contribution is correspondingly suppressed.

For the multi-table version, each $(s,n)$ forms a retrieval branch. To prevent the output scale from becoming unstable as the number of branches increases, we adopt normalized fusion rather than independent sigmoid gating. Specifically, for the same order $n$, all subtables share the same readout projection:
\[
k_{t,s}^{(n)} = W_K^{(n)} e_{t,s}^{(n)} + b_K^{(n)},
\qquad
v_{t,s}^{(n)} = W_V^{(n)} e_{t,s}^{(n)} + b_V^{(n)}.
\]

We define the branch score as
\[
\rho_{t,s}^{(n)} =
\frac{
\mathrm{RMSNorm}(h_t)^\top \mathrm{RMSNorm}(k_{t,s}^{(n)})
}{
\sqrt{d}
},
\]
and apply softmax normalization over all $(s,n)$ branches:
\[
\pi_{t,s}^{(n)} =
\mathrm{softmax}_{(s,n)}\!\left(\rho_{t,s}^{(n)} / \tau_f\right),
\]
where $\tau_f > 0$ is the fusion temperature. The final readout is
\[
v_t =
\sum_{s=1}^{S} \sum_{n\in\mathcal{N}} \pi_{t,s}^{(n)} v_{t,s}^{(n)}.
\]

This design allows different subtables to compete under the same contextual query while keeping the readout space consistent within the same order.

To enlarge the short-range receptive field and enhance nonlinearity, we further introduce a short depthwise-separable causal convolution branch over the fused sequence
\[
V = [v_1,\ldots,v_T]^\top
\]
as
\[
Y = V + \mathrm{SiLU}\!\bigl(\mathrm{DWConv1D}(\mathrm{RMSNorm}(V))\bigr).
\]

Here the convolution kernel size $w$ is set to $4$, and the dilation rate $\delta$ is typically set to $\max \mathcal{N}$. Lngram is integrated into the backbone as a residual branch before attention:
\[
H^{(\ell)} \leftarrow H^{(\ell)} + Y.
\]

The original self-attention and feedforward network of the base layer are then applied as usual.

\subsection{Backpropagation}

Because the discrete symbols are produced by the hard threshold
\[
b_{t,c} = \mathbb{I}[z_{t,c} > 0]
\]
and the table addresses are constructed exactly from these discrete symbols, the true forward path of the retrieval branch with respect to $Z$ is a piecewise constant function whose gradient is zero almost everywhere. Directly applying the straight-through estimator (STE) to the threshold function fails to capture the structured dependency of ``bit $\to$ symbol $\to$ address $\to$ table entry,'' which in our experiments manifests as a failure to converge. To address this issue, we use a counterfactual surrogate gradient. The core idea is that, during backpropagation, we hold other local conditions fixed and explicitly compare the retrieval results corresponding to several counterfactual symbol codes, then use these differences to approximate the effect of the discrete decision on the loss. Importantly, this surrogate is applied only to the discretization projection; table entries, readout projections, and the convolution branch still receive exact gradients under the standard chain rule.

Specifically, for a local route position, we treat its $M$ bit logits as independent Bernoulli variables, enumerate all $K=2^M$ local symbols, and construct the conditional expectation retrieval vector
\[
\begin{aligned}
\mu(z) &= \sum_{c=0}^{K-1} P(c\mid z)\, E_c, \\
P(c\mid z) &=
\prod_{j=0}^{M-1}
p_j^{\beta_j(c)}
(1-p_j)^{1-\beta_j(c)}.
\end{aligned}
\]
where $p_j=\sigma(\tau z_j)$, $\beta_j(c)$ denotes the $j$-th bit of symbol $c$, and $E_c$ denotes the retrieval result corresponding to the associated counterfactual symbol. We then analytically differentiate this local surrogate.

In our experiments, we find that a one-bit approximate surrogate can significantly improve computational efficiency while retaining most of the accuracy. Specifically, for bit $j$, we compare only the two retrieval results $E_j^{(0)}$ and $E_j^{(1)}$ obtained by forcing that bit to $0$ or $1$, respectively, and approximate its gradient by
\[
\frac{\partial \mathcal{L}}{\partial z_j}
\approx
\lambda \tau p_j(1-p_j)
\left\langle g, E_j^{(1)} - E_j^{(0)} \right\rangle
\]
where $g$ is the upstream gradient and $\lambda$ is a global scaling coefficient. A detailed derivation of the gradient is provided in Appendix~\ref{app:surrogate-gradient}.

\section{Experiments}

This section evaluates Lngram from four perspectives: general language understanding after text pretraining, long-context language modeling, post hoc domain knowledge injection, and cross-modal transfer to VLA/VLM. The experiments are designed primarily to test whether, under the same or a controlled parameter budget, delegating local static patterns to a conditional lookup module can improve the effective computational capacity of the Transformer backbone, and whether this mechanism can transfer to pretrained-model adaptation and non-text modalities.

\subsection{Evaluation on General Language Tasks}
\label{sec:nlp_experiment}

We first validate the effectiveness of Lngram on language models trained from scratch. We compare three models at the $2$B scale: MOE, MOE+Engram, and MOE+Lngram. MOE follows the MoE architecture of DeepSeek-V2 \citep{dai2024deepseekmoe,liu2024deepseekv2}, with $1$ shared expert and $16$ routed experts. For MOE+Engram and MOE+Lngram, we reduce the number of routed experts from $16$ to $12$ and allocate the released $\sim 0.5$B parameters to the Engram or Lngram modules, thereby keeping the total parameter count approximately unchanged.

Both Engram and Lngram are inserted into layers $2$ and $12$ of the model, with the maximum N-gram order set to $3$. Lngram uses $4$-bit routes and exact $2/3$-gram table lookup. The memory-table parameters are trained with a higher learning rate and no weight decay, while the convolution branch is zero-initialized to reduce perturbations from the newly introduced branch to the backbone path during the early stage of training. It should be noted that both Engram and Lngram include additional readout projection parameters, so the number of activated parameters per token is about $40$M higher than that of the MOE baseline, accounting for less than $10\%$. We retain this difference under the same total parameter budget. To further control for the possible influence of the increased activated parameters, we additionally introduce MoE+Lngram-23L, whose total and activated parameter counts are both smaller than those of the MoE baseline ($1.9$B / $0.4$B vs. $2$B / $0.42$B), yet it still achieves better performance. This result further strengthens our argument, and a detailed analysis of this model is provided in Section~\ref{sec:layer_reduction}.

All models are trained on $35$B tokens sampled from FineWeb-Edu \citep{penedo2024fineweb}. The backbone is optimized with AdamW, using a peak learning rate of $3 \times 10^{-4}$, weight decay of $0.01$, and gradient clipping threshold of $1.0$. The learning rate follows a cosine decay schedule, and the batch size is $0.5$B tokens. All models use the Llama2 tokenizer \citep{touvron2023llama} with a vocabulary size of $32{,}000$. After training, evaluation is conducted with lm-eval \citep{biderman2024lessons,zellers2019hellaswag,hendrycks2021measuring,welbl2017crowdsourcing,sakaguchi2021winogrande,bisk2020piqa}.

\begin{table*}[t]
    \centering
    \caption{Results on general language tasks.}
    \label{tab:general_language}
    \begin{tabular}{lcccccc}
        \toprule
        Model & HellaSwag & MMLU & SciQ & WinoGrande & PIQA & AVG \\
        \midrule
        MOE & 0.4394 & 0.2435 & 0.6830 & 0.5272 & 0.6801 & 0.5146 \\
        MOE+Engram & 0.4418 & 0.2502 & 0.6980 & 0.5383 & 0.6844 & 0.5225 \\
        MOE+Lngram & \textbf{0.4481} & \textbf{0.2619} & \textbf{0.7100} & 0.5312 & \textbf{0.6926} & \textbf{0.5288} \\
        MOE+Lngram-23L & 0.4475 & 0.2300 & 0.7080 & \textbf{0.5438} & 0.6740 & 0.5207 \\
        \bottomrule
    \end{tabular}
\end{table*}

Table~\ref{tab:general_language} shows that introducing conditional memory improves overall model performance. Among the compared methods, Lngram achieves the highest average score, outperforming the baseline on all tasks and surpassing Engram on four of the five tasks. Compared with the MOE baseline, Lngram improves the average score by 1.41 percentage points; compared with Engram, it further improves by 0.62 percentage points. The gains are concentrated mainly on MMLU, SciQ, and PIQA, suggesting that the discrete symbols learned by Lngram can provide effective local memory support for knowledge-intensive and commonsense judgments.

\paragraph{Statistical Significance.}
We assess significance using paired bootstrap resampling over per-example correctness indicators, treating each evaluation instance as a paired unit. For each model pair, we run $10{,}000$ bootstrap trials, report percentile $95\%$ confidence intervals, and compute two-sided bootstrap $p$-values with Holm--Bonferroni correction across benchmarks.

\begin{table*}[t]
    \centering
    \caption{Paired-bootstrap significance results for Table~\ref{tab:general_language}.}
    \label{tab:nlp_significance}
    \begin{tabular}{lcccc}
        \toprule
        Benchmark & Evaluation $N$ & $\Delta$ & 95\% CI & Adjusted $p$ \\
        \midrule
        HellaSwag & 10{,}042 & +0.87 & $[+0.21,\,+1.55]$ & 0.024 \\
        MMLU & 14{,}042 & +1.84 & $[+1.12,\,+2.57]$ & $<0.001$ \\
        SciQ & 1{,}000 & +2.70 & $[+0.60,\,+4.90]$ & 0.042 \\
        WinoGrande & 1{,}267 & +0.40 & $[-1.89,\,+2.68]$ & 0.614 \\
        PIQA & 1{,}838 & +1.25 & $[+0.30,\,+2.21]$ & 0.040 \\
        \bottomrule
    \end{tabular}
\end{table*}

Table~\ref{tab:nlp_significance} shows that MoE+Lngram improves over MoE on all five benchmarks, with a task-balanced average gain of $+1.41$ percentage points. HellaSwag, MMLU, SciQ, and PIQA remain significant after Holm--Bonferroni correction, while WinoGrande shows a positive but non-significant gain. Overall, the results indicate consistent improvements across benchmarks.

\paragraph{Scaling on NLP Tasks.}
The results above demonstrate the effectiveness of Lngram under the original $35$B-token training setting.
To examine whether this advantage remains stable under larger training budgets or increased model capacity, we conduct two additional scaling experiments.
First, we train the $2$B model for $140$B tokens.
Second, we scale the model to $8$B total parameters with $1.6$B activated parameters, while keeping the training budget at $35$B tokens.
In each setting, we compare MoE and MoE+Lngram under the same corresponding scale.

\begin{table*}[t]
\centering
\caption{Scaling results on NLP tasks.}
\label{tab:nlp_scaling}
\resizebox{\linewidth}{!}{
\begin{tabular}{llccccccc}
\toprule
Scale & Model & ARC-E & ARC-C & SciQ & WinoGrande & HellaSwag & MMLU & AVG \\
\midrule
$2$B / $140$BT & MoE & 0.5471 & 0.3183 & 0.7280 & \textbf{0.5249} & 0.4776 & 0.2346 & 0.4718 \\
$2$B / $140$BT & MoE+Lngram & \textbf{0.5492} & \textbf{0.3208} & \textbf{0.7290} & 0.5209 & \textbf{0.4860} & \textbf{0.2476} & \textbf{0.4756} \\
\midrule
$8$B / $35$BT & MoE & 0.5589 & 0.3353 & 0.7520 & \textbf{0.5272} & 0.4964 & 0.2386 & 0.4847 \\
$8$B / $35$BT & MoE+Lngram & \textbf{0.5804} & \textbf{0.3464} & \textbf{0.7660} & 0.5249 & \textbf{0.5002} & \textbf{0.2518} & \textbf{0.4950} \\
\bottomrule
\end{tabular}
}
\end{table*}

Table~\ref{tab:nlp_scaling} shows that Lngram maintains its advantage under both data and parameter scaling.
When the $2$B model is trained for $140$B tokens, MoE+Lngram improves the average score from $0.4718$ to $0.4756$, corresponding to a gain of $0.38$ percentage points.
Although longer training allows the MoE baseline to better absorb local patterns into its dense parameters, Lngram still outperforms it on five of the six benchmarks.
The gains are particularly clear on MMLU and HellaSwag, where Lngram improves performance by $1.30$ and $0.84$ percentage points, respectively.
These results suggest that the conditional memory branch is not merely useful in early or under-trained regimes.
Even with substantially more training data, explicit latent-space lookup continues to provide complementary information that the dense backbone may not capture or exploit as efficiently.

The benefit of Lngram becomes more pronounced when model capacity is increased.
At the $8$B-total-parameter scale with $1.6$B activated parameters, MoE+Lngram improves the average score from $0.4847$ to $0.4950$, yielding a gain of $1.03$ percentage points.
It outperforms the MoE baseline on ARC-E, ARC-C, SciQ, HellaSwag, and MMLU, with especially large gains of $2.15$ percentage points on ARC-E, $1.40$ on SciQ, and $1.32$ on MMLU.
Compared with the $2$B / $140$BT Lngram model, the $8$B / $35$BT Lngram model further improves the average score by $1.94$ percentage points, despite using fewer training tokens.
This indicates that Lngram can effectively exploit increased sparse model capacity: as the backbone and memory branch grow, the model can form more expressive latent keys, store richer local static patterns, and retrieve them more selectively through the context-aware readout mechanism.

Overall, these results further demonstrate that the advantage of Lngram is stable across both data and parameter scaling.

\subsection{Long-Context Language Modeling}

\begin{figure}[t]
    \centering
    \includegraphics[width=0.47\textwidth]{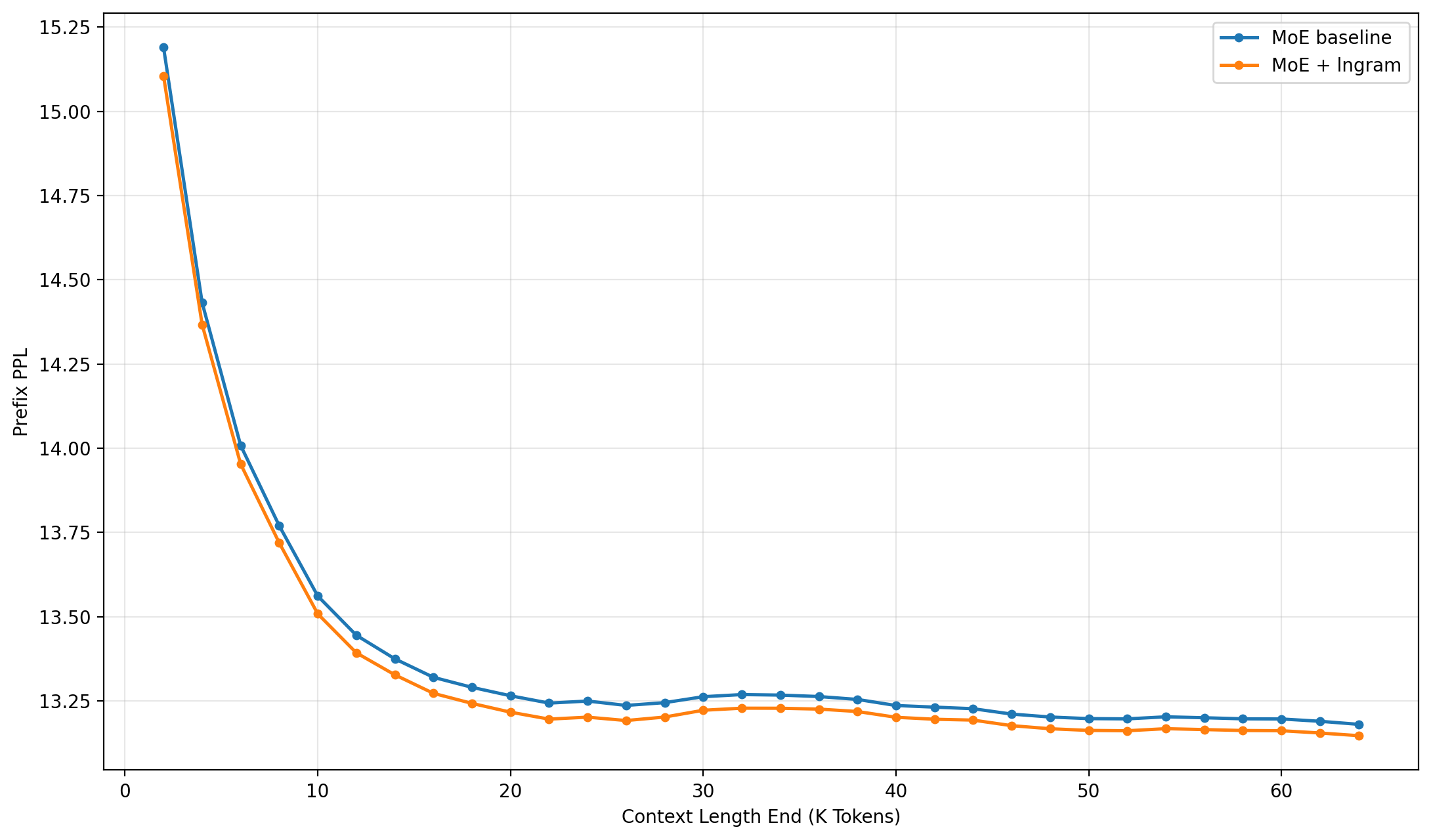}
    \caption{Illustration of prefix perplexity at different context positions in long-context language modeling.}
    \label{fig:long_context_ppl}
\end{figure}

To evaluate the effect of Lngram on long-context language modeling, we perform YaRN-based long-context extension training on MOE and MOE+Lngram \citep{peng2024yarn}. The training set consists of 4,096 samples of 64k-token text drawn from PG-19 \citep{rae2020compressive}, and the test set consists of another non-overlapping 256 samples of 64k-token text. Figure~\ref{fig:long_context_ppl} compares the perplexity (PPL) of the baseline and the Lngram-enhanced model under different prefix lengths.

Figure~\ref{fig:long_context_ppl} shows that the PPL of Lngram remains consistently lower than that of the baseline across the entire test range. The advantage is more pronounced in the short-to-medium prefix regime and persists all the way to 64k. This result is consistent with the design goal of Lngram. Lngram does not replace attention in modeling long-range dependencies; instead, it delegates local static patterns to an exact lookup branch. As a result, the attention and feedforward networks can devote more capacity to cross-segment dependencies and global context integration. The sustained reduction in long-context PPL indicates that the conditional memory branch does not impair the model's ability to exploit long-range context, while providing additional local-pattern priors for long-context language modeling.

\subsection{Lngram-Tuning: Domain Knowledge Injection}

We further test the domain adaptation capability of post hoc Lngram insertion. The base model is Qwen3-1.7B-Base. We add an Lngram module with approximately 200M parameters and train it on about 16 million samples drawn from intelligent driving data, evaluating on two driving-exam datasets, BDD and CNK \citep{zhang2023study}. We compare three settings: Qwen3-1.7b-base+Lngram, where only the newly added Lngram is trained and the base model is frozen; Qwen3-1.7b-base(tuning), where all model parameters are fine-tuned in the standard way; and Qwen3-1.7b-base(tuning)+Lngram, where only Lngram is trained in the first epoch and the entire model is then unfrozen. All methods are trained for 3 epochs on the same domain-specific data. The results are shown in Table~\ref{tab:domain_adaptation}.

\begin{table*}[t]
\centering
\caption{Results of domain knowledge injection.}
\label{tab:domain_adaptation}
\begin{tabular}{lcc}
\hline
Model & BDD acc & CNK acc \\
\hline
Qwen3-1.7b-base & 50.59 & 78.01 \\
Qwen3-1.7b-base+Lngram & 55.73 & 79.39 \\
Qwen3-1.7b-base(tuning) & 56.91 & 79.39 \\
Qwen3-1.7b-base(tuning)+Lngram & \textbf{62.45} & \textbf{81.02} \\
\hline
\end{tabular}
\end{table*}

Training only the newly added Lngram already leads to a substantial improvement in domain performance: BDD increases from 50.59 to 55.73, and CNK from 78.01 to 79.39. This indicates that new knowledge can largely be written into the newly introduced memory branch, without relying on broad modifications to the base model parameters. Furthermore, Qwen3-1.7b-base(tuning)+Lngram achieves the best results on both datasets, outperforming Qwen3-1.7b-base(tuning) by 5.54 and 1.63 percentage points on BDD and CNK, respectively. This suggests that Lngram can not only independently absorb new knowledge, but also provide a better starting point for subsequent joint fine-tuning.

To examine the effect of this domain adaptation on the model's original general capabilities, we further evaluate on HellaSwag, MMLU, PIQA, SciQ, and WinoGrande. The results are shown in Table~\ref{tab:general_drop}. All domain-adaptation methods reduce performance on the original general benchmarks. This suggests that Qwen3-1.7b-base(tuning)+Lngram is better suited as a specialized solution for a target domain, rather than as a cost-free way of appending new knowledge. Its advantage is that, in scenarios with a clearly defined target domain, the domain-specific gains brought by Lngram are substantially larger than those of Qwen3-1.7b-base(tuning) alone.

\begin{table*}[t]
\centering
\caption{Impact of domain adaptation on original general capabilities.}
\label{tab:general_drop}
\begin{tabular}{lcccccc}
\hline
Model & HellaSwag & MMLU & PIQA & SciQ & WinoGrande & AVG (acc) \\
\hline
Qwen3-1.7b-base & \textbf{0.6648} & \textbf{0.6048} & \textbf{0.7568} & \textbf{0.9590} & \textbf{0.6448} & \textbf{0.7260} \\
Qwen3-1.7b-base-tuning & 0.6244 & 0.5700 & 0.7301 & 0.9260 & 0.6306 & 0.6962 \\
Qwen3-1.7b-base+Lngram & 0.5863 & 0.5157 & 0.7078 & 0.9380 & 0.6014 & 0.6698 \\
Qwen3-1.7b-base-tuning+Lngram & 0.5893 & 0.5164 & 0.6921 & 0.9210 & 0.5935 & 0.6625 \\
\hline
\end{tabular}
\end{table*}

\subsection{Vision-Language-Action Experiments}

We further test the cross-modal effectiveness of Lngram in the VLA setting. The experiments adopt the overall architecture of pi0.5 \citep{physicalintelligence2025pi05}, with the action model instantiated as either MOE or MOE+Lngram. Considering that VLA models often need to be deployed on edge devices or in low-latency scenarios, we use the multi-subtable version of Lngram in this experiment. Under a given parameter budget, the number of subtables is chosen to be as close as possible to the single-table entry dimension so as to maximize the readout rank. Both models have a total parameter count of 1B and 0.4B activated parameters, ensuring that the comparison mainly reflects structural differences.

Training and evaluation are conducted on the LIBERO benchmark in the StarVLA library \citep{starvla2026starvla,liu2023libero}. 
In the updated setting, both models are first pretrained for an additional 60{,}000 steps, followed by 100{,}000 steps of training on LIBERO-train. 
They are then evaluated on the Long, Goal, Object, and Spatial subsets of LIBERO-test using success rate. 
The results are shown in Table~\ref{tab:vla_libero}.

\begin{table*}[t]
\centering
\caption{Success rates of MindPI-based VLA models on LIBERO-test.}
\begin{tabular}{lccccc}
\toprule
Model & Long & Goal & Object & Spatial & Avg \\
\midrule
MindPI-MoE & 0.962 & \textbf{0.990} & 0.990 & 0.988 & 0.9825 \\
MindPI-MoE+Lngram & \textbf{0.968} & 0.988 & 0.990 & \textbf{0.994} & \textbf{0.9850} \\
\bottomrule
\end{tabular}
\label{tab:vla_libero}
\end{table*}

These results indicate that Lngram can still provide additional gains in  VLA regime. Together with the text and vision-language experiments, this suggests that the conditional matching mechanism of Lngram does not depend on text-token discreteness and can also operate in action-sequence modeling under visual conditioning.

\subsection{Vision-Language Experiments}

We also evaluate the effectiveness of Lngram on vision-language models. The experiments follow the LLaVA paradigm, using SigLIP2 as the vision encoder, a 3-layer MLP as the projector, and Qwen2.5-0.5b-instruct as the language backbone \citep{liu2023visual,tschannen2025siglip2,qwen2025qwen25}. Because this experiment is built on top of a fixed pretrained LLM and vision encoder, it is not possible to form a strictly parameter-matched comparison that differs only in whether Lngram is added. Accordingly, in this section we use the same backbone and training pipeline and compare only the performance before and after adding Lngram.

VLM training consists of three stages: first, the projector is trained on 1M samples; next, the LLM is unfrozen and trained on an additional 15M samples; finally, the vision encoder is unfrozen and trained for another 5M samples. Lngram is introduced in the second stage, directly acting on the LLM and trained jointly with it. Evaluation is conducted on SeedBench \citep{li2024seedbench}. The results are shown in Table~\ref{tab:vlm_results}.

\begin{table*}[t]
\centering
\caption{Evaluation results on vision-language models.}
\label{tab:vlm_results}
\small
\begin{tabular}{lcc}
\hline
Evaluation Dimension & Qwen2.5-0.5b-instruct & Qwen2.5-0.5b-instruct+Lngram \\
\hline
Visual Reasoning      & 71.0 & \textbf{73.1} \\
Text Understanding    & \textbf{42.9} & \textbf{42.9} \\
Spatial Relation      & \textbf{40.2} & 39.9 \\
Scene Understanding   & 71.8 & \textbf{72.0} \\
Instances Counting    & 50.1 & \textbf{50.5} \\
Instance Location     & 44.8 & \textbf{46.1} \\
Instance Interaction  & \textbf{66.0} & 61.9 \\
Instance Identity     & \textbf{67.4} & \textbf{67.4} \\
Instance Attributes   & 61.0 & \textbf{62.8} \\
Average               & 60.5 & \textbf{61.2} \\
\hline
\end{tabular}
\end{table*}

After adding Lngram, the average score improves from 60.5 to 61.2. The gains are concentrated mainly on Visual Reasoning, Instance Location, and Instance Attributes, with visual reasoning improving by 2.1 percentage points. This indicates that once visual features are projected into the language space, Lngram can still effectively exploit the local stable patterns therein. At the same time, the drop on Instance Interaction is relatively noticeable, suggesting that the current configuration still provides limited benefit for modeling inter-instance relations. Overall, these results show that the effectiveness of Lngram can transfer from pure language models to the vision-language setting.

\section{Analysis}
\label{sec:analysis}

This section analyzes the internal working mechanism of Lngram. We first examine whether Lngram improves the model's effective depth from three perspectives: LogitLens \citep{nostalgebraist2020logitlens}, CKA \citep{kornblith2019similarity}, and layer reduction experiments. We then analyze the core hyperparameters, inference efficiency, and memory overhead. Finally, we visualize the gating behavior to inspect the local triggering patterns of Lngram. Some experimental settings in this section follow Engram.

\subsection{Lngram and Effective Depth}
\label{sec:effective_depth}

Standard Transformers typically require multiple layers of computation to gradually complete local pattern composition and entity disambiguation. For example, for a multi-token entity, the model often first forms local phrase-level or category-level cues in shallow layers, and then converges to a specific entity in later layers. Table~\ref{tab:entity_progress} gives a representative example. Since Lngram explicitly represents such local static patterns as lookup operations, its effect should not be reflected merely as output correction in the final few layers; rather, it should appear as intermediate representations approaching the final predictive state earlier. To verify this, we analyze Lngram using LogitLens, CKA, and layer reduction experiments.

\begin{table*}[t]
\centering
\caption{An example showing how a multi-token entity is progressively formed across intermediate layers. This example illustrates that a standard backbone typically requires multiple layers of composition to converge from local token patterns to a specific entity.}
\label{tab:entity_progress}
\begin{tabular}{cl}
\toprule
Layer & Interpretation of the intermediate representation \\
\midrule
1--2 & Interprets ``Wales'' mainly as a concept related to a region in the United Kingdom \\
3 & Further generalizes ``Wales'' to a concept related to a European country or region \\
4 & Forms the title cue ``Princess of Wales,'' but has not yet identified a specific person \\
5 & Associates the title with the spouse of the Prince of Wales \\
6 & Converges to the specific entity ``Diana, Princess of Wales'' \\
\bottomrule
\end{tabular}
\end{table*}

\subsubsection{LogitLens: Prediction-Relevant Information Emerges Earlier}
\label{sec:logitlens}

We first examine whether Lngram changes when prediction-relevant information appears in the network. For the hidden states $H^{(l)}$ at layer $l$, we project them into the vocabulary space using the final LM head, and compute the KL divergence between this intermediate distribution and the final output distribution:
\begin{equation}
D_{\mathrm{KL}}\!\left(p_{\mathrm{final}}\;\Vert\;p_l\right).
\end{equation}
A smaller KL divergence indicates that the hidden states at that layer are closer to the representation needed for the final prediction.

\begin{figure}[t]
\centering
\includegraphics[width=0.78\linewidth]{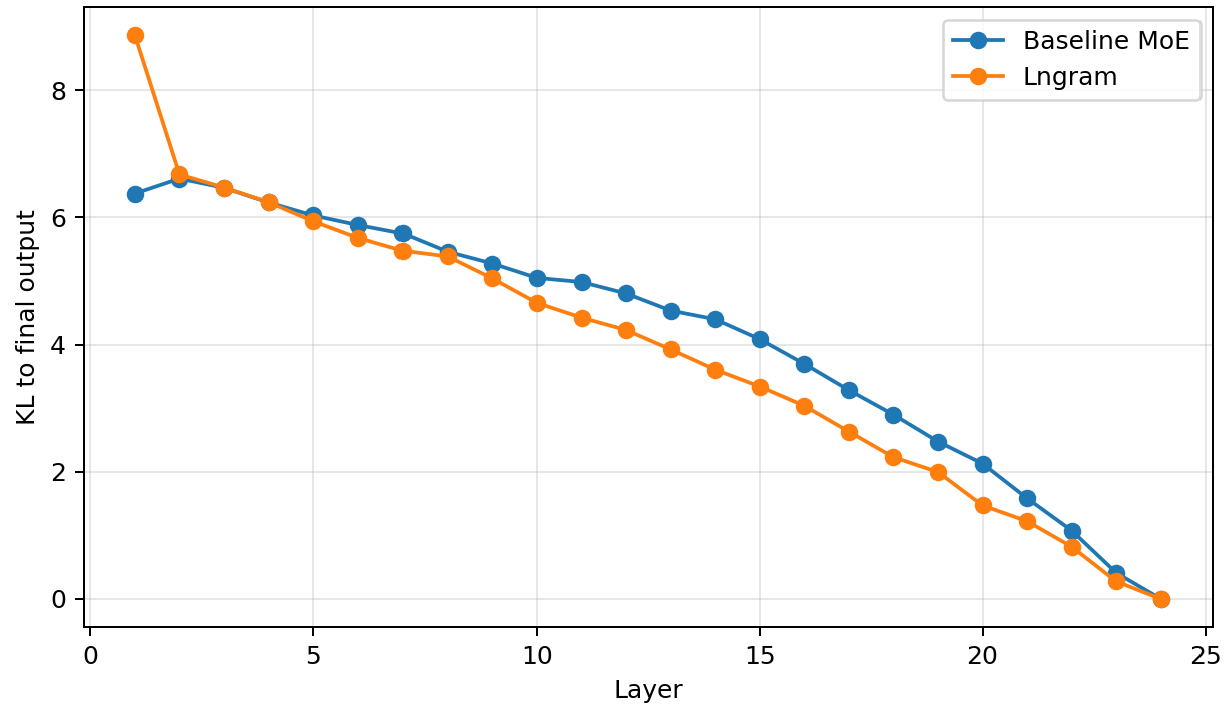}
\caption{LogitLens KL divergence as a function of layer depth. Except for the shallowest layer, where Lngram has not yet been introduced, Lngram exhibits lower KL divergence on most layers.}
\label{fig:logitlens_kl}
\end{figure}

The results are shown in Figure~\ref{fig:logitlens_kl}. Lngram exhibits a noticeable adaptation phase at layer 1, where its KL divergence is higher than that of the baseline. After that, however, the KL divergence decreases more rapidly and remains lower than that of the MoE baseline across most middle and later layers. This indicates that Lngram does not merely correct outputs at the end of the network; instead, it allows some prediction-relevant information to enter the backbone representations earlier. In other words, with the same number of layers, the model with Lngram reaches representational states closer to the final decision at an earlier stage.

\subsubsection{CKA: Representations Align with Deeper Layers Earlier}
\label{sec:cka}

We further use linear CKA to compare the layer representations of the MoE baseline and MoE+Lngram. Given two sets of representations $X$ and $Y$, their linear-kernel Gram matrices are $K=XX^\top$ and $L=YY^\top$, respectively. CKA is defined as
\begin{equation}
\mathrm{CKA}(K,L)
=
\frac{\mathrm{HSIC}(K,L)}
{\sqrt{\mathrm{HSIC}(K,K)\mathrm{HSIC}(L,L)}}.
\end{equation}

To quantify the effective depth corresponding to each Lngram layer, let
$s_{ij}=\mathrm{CKA}(H^{\mathrm{base}}_i,H^{\mathrm{lngram}}_j)$.
For the $j$-th Lngram layer, let $\mathcal{T}_k(j)$ denote the set of the top-$k$ baseline layers with the highest similarity, and define the soft alignment as
\begin{equation}
a_j
=
\frac{\sum_{i\in\mathcal{T}_k(j)} i\,s_{ij}}
{\sum_{i\in\mathcal{T}_k(j)} s_{ij}}.
\end{equation}
The corresponding effective depth gain is
\begin{equation}
\Delta_j=a_j-j.
\end{equation}
If $\Delta_j>0$, this means that the $j$-th Lngram layer is closer to a deeper-layer representation in the baseline model.

\begin{figure}[t]
\centering
\includegraphics[width=0.72\linewidth]{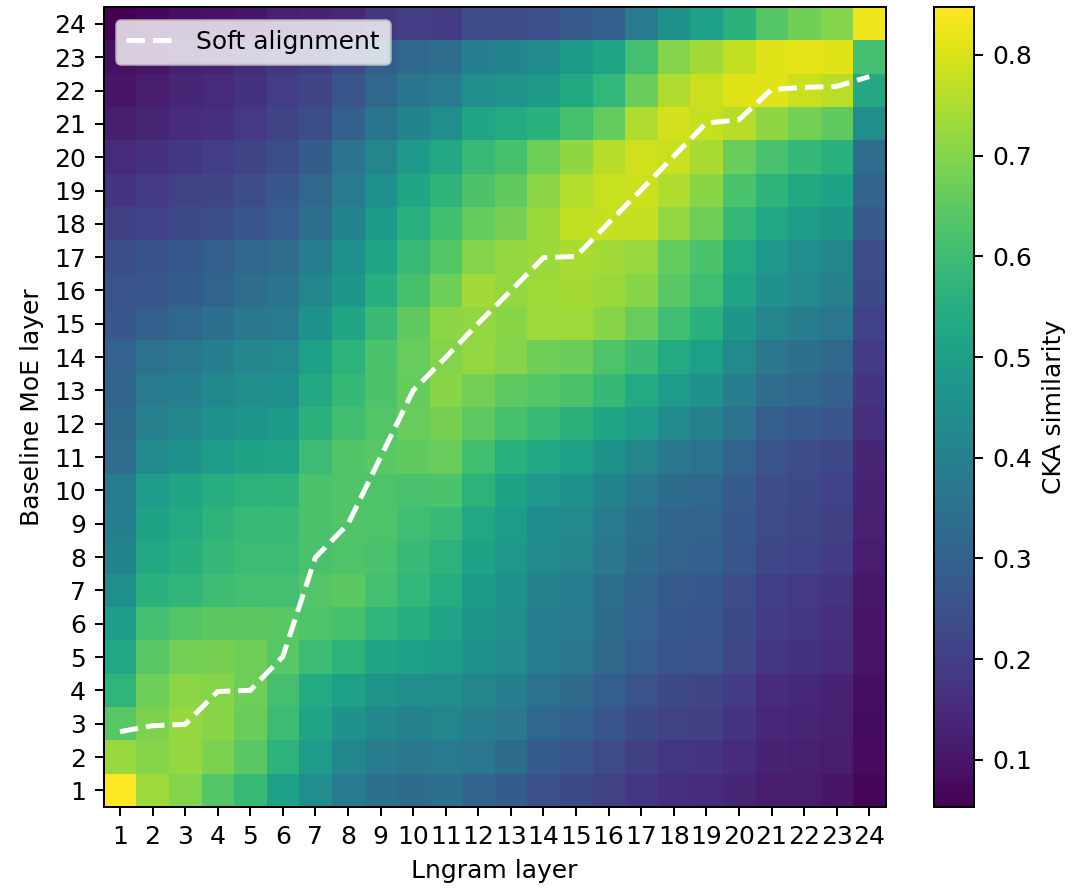}
\caption{CKA similarity between the layer representations of the MoE baseline and MoE+Lngram. The dashed line indicates the soft alignment. The high-similarity region is shifted overall above the diagonal, indicating that Lngram layers are closer to deeper representations in the baseline model.}
\label{fig:cka_heatmap}
\end{figure}

Figure~\ref{fig:cka_heatmap} shows that the high-similarity region in the CKA heatmap is shifted overall above the diagonal, and the soft alignment curve also lies above the same-layer diagonal for most layers. This phenomenon is most pronounced in the middle layers, where the effective depth gain is approximately $+2$ to $+3$ layers for multiple layers. Together with the LogitLens results, this shows that Lngram not only causes prediction-relevant information to emerge earlier, but also gives these representations deeper-layer structure at an earlier stage. Therefore, the benefit of Lngram is better characterized as an increase in effective depth rather than a simple enhancement of late-stage output correction.

\subsubsection{Layer Reduction Validation}
\label{sec:layer_reduction}

To test whether this effective depth gain can partially substitute for backbone depth, under the setting in Section~\ref{sec:nlp_experiment}, we reduce the backbone of MoE+Lngram from 24 layers to 23 layers while keeping all other training settings unchanged. The results are shown in Table~\ref{tab:general_language}.

Despite having one fewer Transformer block, MOE+Lngram-23L still achieves a higher average score than the 24-layer MoE baseline, and obtains better results on HellaSwag, SciQ, and WinoGrande. This indicates that explicit local memory can reduce the backbone depth consumed by local static matching. However, MMLU and PIQA still decline, suggesting that Lngram cannot fully replace general deep computation. More precisely, Lngram provides a structural substitute for modeling a subset of local patterns, rather than a fully equivalent replacement for Transformer depth. In addition, both the total parameter count and the number of activated parameters of MOE+Lngram-23L are smaller than those of the MoE baseline, which further reinforces our argument in Section~\ref{sec:nlp_experiment}.

\begin{figure*}[t]
\centering
\includegraphics[width=0.9\linewidth]{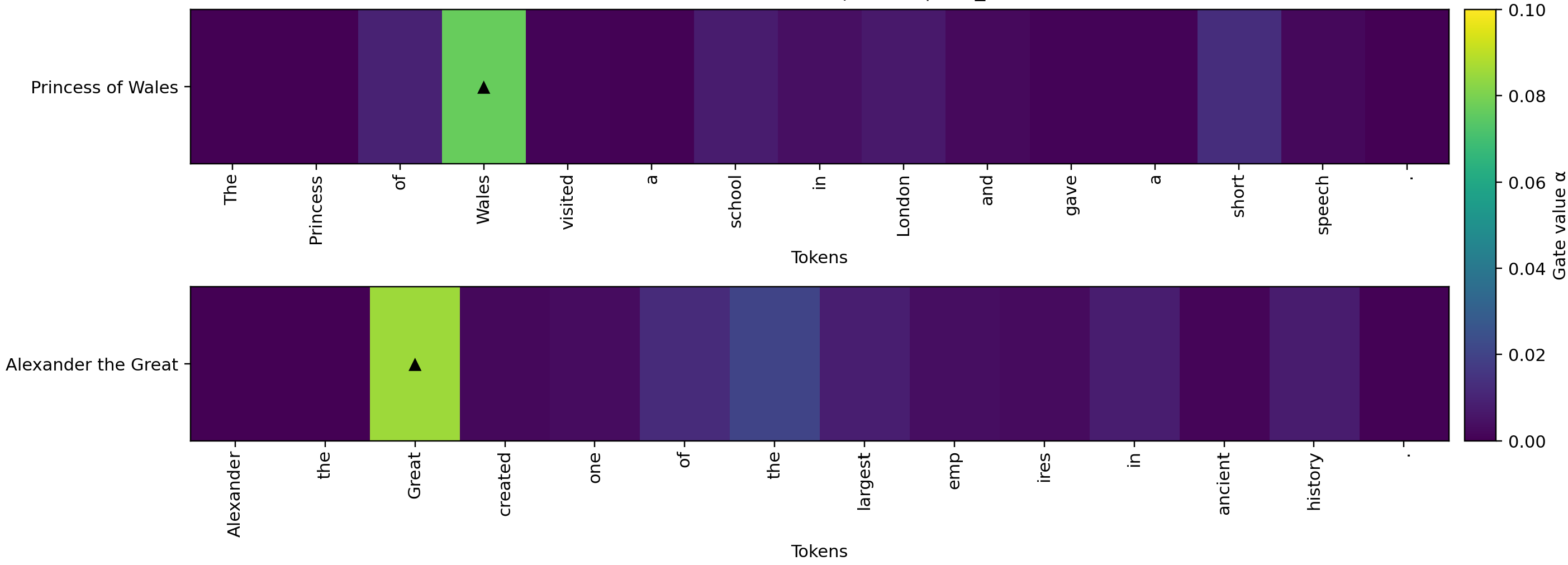}
\caption{Gating responses of the layer-1 3-gram branch in Lngram. Peaks occur where fixed phrases such as ``Great'' and ``Wales'' are completed.}
\label{fig:gate_visualization}
\end{figure*}

\subsection{Ablation on Core Hyperparameters}
\label{sec:ablation}

All ablation experiments in this section use the approximate surrogate gradient and are trained on 30B tokens. Therefore, the goal here is primarily to compare the relative trends under different settings.

\subsubsection{$N$-gram Order Combinations}
\label{sec:ngram_ablation}

We first examine the effect of different $N$-gram order combinations on performance. The results are shown in Table~\ref{tab:prefix_ablation}.

\begin{table*}[t]
\centering
\caption{Ablation on $N$-gram order combinations.}
\label{tab:prefix_ablation}
\begin{tabular}{lccccc}
\toprule
Order Combination & HellaSwag & MMLU & SciQ & WinoGrande & AVG \\
\midrule
$\{2,3\}$ & \textbf{0.4471} & 0.2567 & 0.6990 & \textbf{0.5288} & \textbf{0.4829} \\
$\{3\}$ & 0.4445 & 0.2399 & \textbf{0.7070} & 0.5280 & 0.4799 \\
$\{1,2,3\}$ & 0.4459 & 0.2431 & 0.6900 & 0.4996 & 0.4697 \\
$\{2\}$ & 0.4444 & \textbf{0.2592} & 0.6540 & 0.5217 & 0.4698 \\
\bottomrule
\end{tabular}
\end{table*}

Using 2-gram and 3-gram jointly yields the best average performance. Using only 3-gram is close to optimal on some tasks, but is slightly worse overall. Using only 2-gram, or adding 1-gram on top of that, both reduces the average score. This suggests that the most effective matching scale for Lngram lies mainly in short multi-token patterns. 1-gram matching is too coarse and tends to introduce memories that are only weakly related to the context, whereas using only 3-gram misses shorter but still stable local patterns. The combination of 2-gram and 3-gram achieves a better balance between coverage and matching precision.

\subsubsection{Sparse Capacity Allocation Between Lngram and MoE}
\label{sec:sparse_capacity}

Both Lngram and MoE experts belong to conditional computation or storage structures, but they serve different functions. MoE-MLP is better suited for dynamic nonlinear transformations, whereas Lngram is better suited for retrieving local static patterns. Based on this distinction, we examine the allocation of sparse capacity between them. In Table~\ref{tab:sparse_capacity}, the ratio denotes the share of sparse parameters allocated to MoE-MLP; thus, 75\% corresponds to Lngram accounting for approximately 25\% of the total sparse parameters.

\begin{table*}[t]
\centering
\caption{Ablation on sparse capacity allocation between MoE-MLP and Lngram.}
\label{tab:sparse_capacity}
\begin{tabular}{lccccc}
\toprule
MoE-MLP Ratio & HellaSwag & MMLU & SciQ & WinoGrande & AVG \\
\midrule
75\% & \textbf{0.4471} & 0.2567 & 0.6990 & \textbf{0.5288} & \textbf{0.4829} \\
65\% & 0.4455 & 0.2426 & \textbf{0.7090} & 0.5075 & 0.4762 \\
85\% & 0.4435 & \textbf{0.2599} & 0.6840 & 0.5185 & 0.4765 \\
\bottomrule
\end{tabular}
\end{table*}

The best results are obtained when MoE-MLP accounts for 75\% of the sparse capacity, i.e., when Lngram accounts for approximately 25\% of the total sparse parameters. When Lngram has too little capacity, the memory branch is insufficient to support stable pattern retrieval. When Lngram has too much capacity, dynamic expert computation is squeezed, and overall performance also declines. This suggests that the role of Lngram is not to replace most of the computation performed by MoE, but to take over the subset of local static patterns that are better handled through lookup.

\subsection{Inference Speed and Memory Footprint}
\label{sec:efficiency}

Using the same model scale as in Section~\ref{sec:nlp_experiment}, we measure the inference speed and memory footprint of MoE and MoE+Lngram on a single H200 GPU. The results are shown in Table~\ref{tab:efficiency}.

\begin{table}[t]
\centering
\caption{Inference efficiency and memory footprint of MoE and MoE+Lngram.}
\label{tab:efficiency}
\resizebox{\linewidth}{!}{
\begin{tabular}{lccc}
\toprule
Metric & Lngram & MoE & Lngram / MoE \\
\midrule
Prefill Throughput (tok/s) & 22547.31 & 19314.50 & 1.1674$\times$ \\
Prefill Latency (ms) & 45.416 & 53.017 & 0.8566$\times$ \\
Prefill Peak Memory (MB) & 8158.82 & 7614.11 & 1.0715$\times$ \\
Decode Throughput (tok/s) & 43.00 & 45.88 & 0.9373$\times$ \\
Decode Latency (ms/token) & 23.256 & 21.798 & 1.0669$\times$ \\
Decode Peak Memory (MB) & 7713.60 & 7597.89 & 1.0152$\times$ \\
Decode Peak Incremental Memory (MB) & 7.79 & 7.79 & 1.0000$\times$ \\
\bottomrule
\end{tabular}
}
\end{table}

The resident GPU memory overhead of Lngram is small: both the theoretical model memory footprint and the allocated memory after model instantiation increase by only about 1.4\%. During the prefill stage, MoE+Lngram achieves higher throughput and lower latency than the baseline, indicating that the lookup and readout branch does not become a bottleneck in long-sequence parallel computation. Peak prefill memory increases by about 7.1\%, mainly due to the additional intermediate readout tensors.

During the decode stage, Lngram is slightly slower than the baseline, with latency increasing by about 6.7\%. This is because, under single-token decoding, the lookup, gating, and readout operations can no longer be amortized as effectively by long-sequence parallelism. At the same time, peak decode memory increases by only about 1.5\%, and the peak incremental memory is identical to that of the baseline. This indicates that Lngram does not introduce any new cache structure that grows with generation length; its additional cost comes mainly from static parameters and a small amount of online computation.

\subsection{Case Study: Gate Visualization}
\label{sec:gate_visualization}

Finally, we visualize the gating scalars of Lngram. Figure~\ref{fig:gate_visualization} shows the gating responses of the layer-1 3-gram branch for two examples.

In the two examples ``Alexander the Great'' and ``The Princess of Wales,'' the strongest gating responses occur at the positions of ``Great'' and ``Wales,'' respectively, with peak values of approximately 0.0853 and 0.0766. This indicates that the readout strength of Lngram is not distributed uniformly across the sequence, but instead tends to rise when a local phrase is completed. In other words, when the current token and the preceding context together form a stable multi-token pattern, the model is more likely to invoke static memory.

At the same time, the gating values remain within a relatively small range overall, behaving as continuous modulation rather than a binary switch. This is consistent with the design of Lngram: the retrieved results do not directly replace the backbone representations, but instead participate in subsequent computation as context-controlled residual signals.

Taken together, the results in this section suggest that the main effect of Lngram can be summarized as a reallocation of computational function. It rewrites part of the local static matching originally carried out by multi-layer dense networks into explicit lookup, allowing the backbone network to form deeper representations earlier. At the same time, this benefit depends primarily on short 2/3-gram patterns and a moderate sparse-capacity allocation, while introducing only modest memory and decode-latency overhead during inference.

\section{Conclusion}

This paper proposes Lngram, a conditional memory mechanism that performs exact N-gram conditional matching in latent space. Lngram learns discrete symbols directly from hidden states and performs retrieval and readout over these symbols, thereby rewriting part of the local static pattern matching in Transformers from dense computation into explicit lookup. Compared with N-gram memory built on tokenizer sequences, Lngram is not constrained by tokenization boundaries, uses learnable retrieval keys, and is more readily extendable to non-text modalities.

Experimental results show that Lngram delivers consistent gains across multiple settings. On language models trained from scratch, Lngram outperforms the baseline on all evaluation items and surpasses Engram on most tasks. In long-context language modeling, Lngram maintains lower perplexity throughout the entire test range. In post hoc domain adaptation, training only the newly added Lngram yields domain performance close to that of full fine-tuning, while joint training with the base model further outperforms standard full fine-tuning. On vision-language and vision-language-action models, Lngram also improves average performance. These results indicate that the advantage of Lngram comes from the functional separation between local static matching and backbone computation, rather than from any particular task or training recipe.

Our analysis further shows that Lngram allows prediction-relevant information to enter intermediate representations earlier, and causes multiple layers of representation to align earlier with deeper states in the baseline model. This indicates that Lngram reduces the backbone depth consumed by local static matching, increases effective depth, and leaves more computation available for global context modeling and compositional reasoning. At the same time, the additional system cost of Lngram is small. Its online computation involves only a small number of linear layers, table lookups, and lightweight readout operations, and it introduces no new cache structure that grows with generation length, thereby maintaining strong deployability in terms of inference speed and memory footprint.

Overall, the core value of Lngram lies in reallocating the computational function of Transformers: it explicitly separates from the backbone network the local static retrieval operations that are better implemented by lookup. Our results show that this approach can improve the efficiency with which models use parameters and depth at low system cost, while providing a scalable conditional memory primitive for language and multimodal models.

\bibliography{refs}
\bibliographystyle{icml2026}

\newpage
\appendix
\onecolumn

\section{Detailed Derivation of the Counterfactual Surrogate Gradient}
\label{app:surrogate-gradient}

\subsection{Notation and Problem Setup}

Consider a fixed $n$-gram order $n$, a fixed route $r$, and a local position $u\in\{0,\ldots,n-1\}$ within the $n$-gram ending at position $t$. Let the bit logits at this local position be
\[
z=(z_0,\ldots,z_{M-1}) \in \mathbb{R}^M,
\qquad
p_j = \sigma(\tau z_j).
\]
Here $\tau>0$ is the temperature coefficient.

When constructing the surrogate, we fix all local symbols in the $n$-gram except the current position, and also fix the symbols of all other routes. For any local symbol
\[
c\in\{0,\ldots,K-1\},
\qquad
K=2^M,
\]
let $\beta(c)\in\{0,1\}^M$ denote its binary representation, and let $\beta_j(c)$ denote its $j$-th bit. After replacing the current position with symbol $c$, the retrieved embedding is denoted by
\[
E_c \in \mathbb{R}^{d_m}.
\]

Let the upstream gradient be
\[
g = \frac{\partial \mathcal{L}}{\partial E} \in \mathbb{R}^{d_m}.
\]
Here $E$ denotes the slice of the retrieval result corresponding to the current position. In the full-lookup and streaming implementations, the source of $g$ differs, but the form of the local surrogate is the same.

\subsection{Local Exact Surrogate}

Here, ``exact'' refers to the analytic gradient of the local expected surrogate, rather than the gradient of the original hard lookup path. We define the local conditional expected retrieval vector as
\[
\mu(z) = \sum_{c=0}^{K-1} P(c\mid z)\, E_c,
\]
where
\[
P(c\mid z)=
\prod_{j=0}^{M-1}
p_j^{\beta_j(c)}
(1-p_j)^{1-\beta_j(c)}.
\]

This corresponds to the probability that the local symbol takes the value $c$ when the $M$ bits at the current position are sampled independently.

We define the local surrogate objective as
\[
\mathcal{L}_{\mathrm{local}}(z)
=
\langle g,\mu(z)\rangle
=
\sum_{c=0}^{K-1} P(c\mid z)\,\langle g,E_c\rangle.
\]

To compute its derivative with respect to $z_j$, we first write
\[
\log P(c\mid z)
=
\sum_{j=0}^{M-1}
\Bigl(
\beta_j(c)\log p_j
+
(1-\beta_j(c))\log(1-p_j)
\Bigr).
\]

Since $p_j=\sigma(\tau z_j)$, we have
\[
\frac{\partial p_j}{\partial z_j}
=
\tau p_j(1-p_j).
\]

It then follows that
\[
\frac{\partial \log P(c\mid z)}{\partial z_j}
=
\tau\bigl(\beta_j(c)-p_j\bigr),
\]
and therefore
\[
\begin{aligned}
\frac{\partial P(c\mid z)}{\partial z_j}
&= P(c\mid z)\,
\frac{\partial \log P(c\mid z)}{\partial z_j} \\
&= \tau P(c\mid z)\bigl(\beta_j(c)-p_j\bigr).
\end{aligned}
\]

Substituting this into the definition of $\mathcal{L}_{\mathrm{local}}$, we obtain
\[
\begin{aligned}
\frac{\partial \mathcal{L}_{\mathrm{local}}}{\partial z_j}
&=
\sum_{c=0}^{K-1}
\frac{\partial P(c\mid z)}{\partial z_j}
\langle g,E_c\rangle \\
&=
\tau \sum_{c=0}^{K-1}
P(c\mid z)\bigl(\beta_j(c)-p_j\bigr)\langle g,E_c\rangle.
\end{aligned}
\]

This is the analytic form of the local exact surrogate. In implementation, summing the above local contributions over all valid positions, all routes, and all $n$-gram orders yields the surrogate gradient for the routing logits. When $M=4$, we have $K=16$, so the cost of enumerating all local symbols remains manageable, making this approach practical in practice.

\subsection{One-Bit Approximate Surrogate}

When computational efficiency is prioritized, one may consider only the local counterfactuals of a single bit. Let $\hat{c}$ denote the hard symbol in the forward pass at the current position, and let its $j$-th bit be $\hat{b}_j$. Let
\[
\hat{c}_j^{(0)}, \qquad \hat{c}_j^{(1)}
\]
denote the two counterfactual symbols obtained by forcing the $j$-th bit to $0$ or $1$, respectively, while keeping all other bits unchanged. The corresponding retrieval results are denoted by
\[
E_j^{(0)}, \qquad E_j^{(1)}.
\]

We define the local counterfactual score for this bit as
\[
s_j
=
\left\langle g, E_j^{(1)} - E_j^{(0)} \right\rangle.
\]

Its meaning is: if only the $j$-th bit is changed, how much does the retrieved result change along the direction of the upstream gradient?

Using the sigmoid slope to characterize the local sensitivity of this bit to the continuous logits, we obtain the one-bit approximate surrogate
\[
\frac{\partial \mathcal{L}_{\mathrm{local}}}{\partial z_j}
\approx
\lambda \tau p_j(1-p_j)s_j
=
\lambda \tau p_j(1-p_j)
\left\langle g, E_j^{(1)} - E_j^{(0)} \right\rangle,
\]
where $\lambda$ is a global scaling coefficient. Compared with the local exact surrogate, this approximation requires constructing only two counterfactual retrieval results for each bit, thus reducing the computational cost from $O(K)$ to $O(M)$.

\subsection{Relation to the full-lookup / streaming Implementations}

The derivations above are all given for a single local position. To apply them to a concrete implementation, one only needs to specify the sources of $g$ and the counterfactual embeddings $E_c$.

In the full-lookup implementation, the retrieval results of all orders are first flattened into a tensor of shape
\[
[B,T,|\mathcal{N}|\cdot R\cdot d_m].
\]
During surrogate backpropagation, the gradient slice corresponding to the current position, current order, and current route is extracted from this flattened tensor as $g$, and the corresponding table entry is retrieved via the counterfactual address as $E_c$.

In the streaming implementation, retrieval and projection are performed in blocks of $(n,\text{route-chunk})$. In this case, $g$ comes from the upstream gradient slice corresponding to the current block, and $E_c$ comes from the counterfactual lookup results within that block. The local formula is exactly the same as in full-lookup; the only difference is that the computation is carried out over smaller blocks in multiple passes, and the results are finally accumulated into the overall gradient of the routing logits.

Regardless of which implementation is used, the table lookup, readout projections, and convolution branch on the main path all receive gradients under the standard chain rule; only the gradient of the routing logits is replaced by the counterfactual surrogate described above. The derivation for multi-table Lngram is exactly the same: one only needs to interpret $E_c$ as the local retrieval result on the corresponding subtable branch.

\section{Verifying Information Retention in Binary Discretized Hidden States}
\label{app:binary_hidden_state}

This appendix verifies a basic assumption made in the introduction: after learnable projection and binary discretization, the hidden states still retain sufficient task-relevant information to serve as keys for conditional matching and retrieval. What is being tested here is information retention at the functional level, namely whether the binary symbols are sufficient to support effective matching; it does not require the binary codes to reconstruct the original continuous vectors losslessly.

\subsection{Verification Method}

Given hidden states at a certain layer, $H\in\mathbb{R}^{T\times d}$, we first apply normalization and linear projection:
\begin{equation}
U=\mathrm{LN}(H),\qquad Z=UW,
\end{equation}
and then perform hard-threshold binarization on each dimension:
\begin{equation}
b_{t,c}=\mathbb{I}[z_{t,c}>0].
\end{equation}
Consistent with the Lngram setting in the main text, we partition every $M$ bits into one route and pack them into a discrete symbol:
\begin{equation}
a_{t,r}=\sum_{j=0}^{M-1} b_{t,(r,j)}2^j .
\end{equation}
In the experiment, we set $M=4$, so the symbol vocabulary size of each route is $2^4=16$.

To test whether these binary symbols retain sufficient information, we construct a long-context setting: we replace the model's global attention with windowed attention, and additionally introduce a retrieval branch that performs addressing based only on the binary symbols above. This branch is not allowed to use the original continuous hidden states directly as retrieval keys. If the binary symbols had lost most of the contextual information, this branch would not be able to effectively recover the performance of windowed attention on long-context tasks. The experimental base model is Qwen3-1.7B, with a window size of 2048. The training process includes alignment training for the newly introduced branch, continued long-context training, and supervised fine-tuning. Evaluation covers both general language understanding tasks and long-context tasks.

\subsection{Results on General Capabilities}

Table~\ref{tab:binary_general_eval} reports the results on general language understanding tasks. After adding the binary-state retrieval branch, the model achieves an average score of 0.7050, close to the 0.7100 of the global-attention baseline. The differences on individual tasks are also small, indicating that the binary discretization branch does not significantly impair the model's existing general capabilities.

\begin{table*}[t]
\centering
\caption{Results of the binary-state retrieval model on general language understanding tasks.}
\label{tab:binary_general_eval}
\resizebox{\linewidth}{!}{
\begin{tabular}{lccccccc}
\toprule
Model & HellaSwag & LAMBADA-OAI & MMLU & PIQA & SciQ & WinoGrande & AVG \\
\midrule
Global-Attn & \textbf{0.6648} & \textbf{0.6295} & \textbf{0.6048} & \textbf{0.7568} & \textbf{0.9590} & \textbf{0.6448} & \textbf{0.7100} \\
Window-Attn + Binary-State Retrieval & 0.6558 & 0.6256 & 0.6033 & 0.7519 & 0.9540 & 0.6393 & 0.7050 \\
\bottomrule
\end{tabular}
}
\end{table*}

\subsection{Long-Context Results}

The long-context results are shown in Table~\ref{tab:binary_longbench}. With windowed attention alone, the model's average score drops from 59.21 under global attention to 29.41, indicating that windowed attention cannot cover the long-range information required by many tasks. After adding the binary-state retrieval branch, the average score recovers to 57.14, reaching about 96.5\% of the global-attention baseline and recovering about 93.1\% of the performance lost by windowed attention relative to global attention. On NIAH-32k, windowed attention alone achieves only 6.20, whereas the binary-state retrieval model recovers to 100.00, matching global attention.

\begin{table*}[t]
\centering
\caption{Results of the binary-state retrieval model on long-context tasks.}
\label{tab:binary_longbench}
\resizebox{\linewidth}{!}{
\begin{tabular}{lccccccc}
\toprule
Model & SAMSum & TriviaQA & MultiNews & TREC & GovReport & NIAH-32k & AVG \\
\midrule
Global-Attn & \textbf{42.04} & \textbf{86.20} & 23.23 & \textbf{72.67} & \textbf{31.11} & \textbf{100.00} & \textbf{59.21} \\
Window-Attn & 32.51 & 61.56 & 10.43 & 52.67 & 13.08 & 6.20 & 29.41 \\
Window-Attn + Binary-State Retrieval & 40.53 & 84.34 & \textbf{23.76} & 68.00 & 26.19 & \textbf{100.00} & 57.14 \\
\bottomrule
\end{tabular}
}
\end{table*}

\subsection{Conclusion}

The above results show that binary discretized hidden states do not lose critical task-relevant information. Although each continuous dimension is compressed into only one bit, the multi-route binary symbols still provide effective matching keys, enabling the model to locate and exploit information relevant to the current context. This supports the discretization design of Lngram: rather than relying on tokenizer IDs to construct keys, Lngram can learn binary symbols directly from hidden states and perform exact local conditional matching over these symbols.

\section{Detailed Model Architecture and Training Hyperparameters}
\label{app:hparams}

This section presents the detailed model architecture and training hyperparameters for MOE, MOE+Engram, and MOE+Lngram, as shown in Table~\ref{tab:appendix_hparams}.

\begin{table*}[t]
\centering
\footnotesize
\setlength{\tabcolsep}{4pt}
\renewcommand{\arraystretch}{1.0}
\caption{Key hyperparameter settings for the main experiments}
\label{tab:appendix_hparams}

\begin{tabular}{lccc}
\toprule
Hyperparameter & MoE & MoE+Engram & MoE+Lngram \\
\midrule
\multicolumn{4}{l}{\textbf{Shared Backbone Settings}} \\
hidden size / layers & \multicolumn{3}{c}{1024 / 24} \\
heads / kv heads & \multicolumn{3}{c}{16 / 2} \\
expert intermediate size & \multicolumn{3}{c}{1536} \\
seed & \multicolumn{3}{c}{42} \\
\midrule
\multicolumn{4}{l}{\textbf{MoE Settings}} \\
routed / shared experts & 16 / 1 & 12 / 1 & 12 / 1 \\
experts per token & \multicolumn{3}{c}{2} \\
\midrule
\multicolumn{4}{l}{\textbf{Training Settings}} \\
backbone optimizer & \multicolumn{3}{c}{AdamW} \\
base learning rate & \multicolumn{3}{c}{$3\times10^{-4}$} \\
backbone weight decay & \multicolumn{3}{c}{0.01} \\
scheduler / warmup ratio & \multicolumn{3}{c}{cosine / 0.01} \\
bf16 & \multicolumn{3}{c}{True} \\
batch & \multicolumn{3}{c}{0.5B tokens} \\
\midrule
\multicolumn{4}{l}{\textbf{Memory Module Settings}} \\
inserted layers & -- & $\{2,12\}$ & $\{2,12\}$ \\
key source & -- & tokenizer ids & hidden-state bits \\
max $n$-gram order & -- & $\{2,3\}$ & $\{2,3\}$ \\
memory dim & -- & 512 & 144 \\
conv kernel size & -- & 4 & 4 \\
\midrule
\multicolumn{4}{l}{\textbf{Engram-Specific}} \\
tokenizer compression & -- & True & -- \\
hash heads & -- & 8 & -- \\
vocab size per table & -- & 524288 & -- \\
\midrule
\multicolumn{4}{l}{\textbf{Lngram-Specific}} \\
bits per route & -- & -- & 4 \\
$n$-gram orders & -- & -- & $\{2,3\}$ \\
surrogate gradient & -- & -- & True \\
surrogate temp / scale & -- & -- & 1.0 / 1.0 \\
\midrule
\multicolumn{4}{l}{\textbf{Memory Table Optimization Rules}} \\
lr multiplier / weight decay & \multicolumn{3}{c}{$5\times$ / 0.0} \\
optimizer & \multicolumn{3}{c}{Adam} \\
\bottomrule
\end{tabular}
\end{table*}

\end{document}